\documentclass[letterpaper, 10 pt, conference]{ieeeconf}  

\IEEEoverridecommandlockouts                              

\usepackage{graphics} 
\usepackage{graphicx}
\usepackage{url}
\usepackage{amsmath}
\usepackage{amsfonts}
\usepackage{xcolor}
\usepackage{listings}
\usepackage{subcaption}

\title{\LARGE \bf
    A software toolkit and hardware platform for investigating and comparing robot autonomy algorithms in simulation and reality
}

\author{Asher Elmquist$^{1}$, Aaron Young$^{2}$, Ishaan Mahajan$^{2}$, Kyle Fahey$^{2}$, Abhiraj Dashora$^{2}$\\
	Sriram Ashokkumar$^{2}$, Stefan Caldararu$^{2}$, Victor Freire$^{3}$, Xiangru Xu$^{4}$, Radu Serban$^{5}$, and Dan Negrut$^{6}$
\thanks{$^{1}$Ph.D. student with the Department of Mechanical Engineering at the University of Wisconsin-Madison, Madison WI, USA
	{\tt\small amelmquist@wisc.edu}}%
\thanks{$^{2}$Undergraduate Student with the Simulation-Based Engineering Lab at the University of Wisconsin-Madison, WI, USA}%
\thanks{$^{3}$Masters student with the Department of Mechanical Engineering at the University of Wisconsin-Madison, Madison WI, USA
        {\tt\small freiremelgiz@wisc.edu}}%
\thanks{$^{4}$Assistant Professor in the Department of Mechanical Engineering at the University of Wisconsin-Madison, WI, USA
        {\tt\small xiangru.xu@wisc.edu}}%
\thanks{$^{5}$Scientist in the Department of Mechanical Engineering at the University of Wisconsin-Madison, Madison WI, USA
    	{\tt\small serban@wisc.edu}}%
\thanks{$^{6}$Professor in the Department of Mechanical Engineering at the University of Wisconsin-Madison, Madison WI, USA
	{\tt\small negrut@wisc.edu}}%
}

\begin{document}

\maketitle
\thispagestyle{empty}
\pagestyle{empty}

\begin{abstract}	
	We describe a software framework and a hardware platform used in tandem for the design and analysis of robot autonomy algorithms in simulation and reality. The software, which is open source, containerized, and operating system (OS) independent, has three main components: a ROS 2 interface to a C++ vehicle simulation framework (Chrono), which provides high-fidelity wheeled/tracked vehicle and sensor simulation; a basic ROS 2-based autonomy stack for algorithm design and testing; and, a development ecosystem which enables visualization, and hardware-in-the-loop experimentation in perception, state estimation, path planning, and controls. The accompanying hardware platform is a 1/6th scale vehicle augmented with reconfigurable mountings for computing, sensing, and tracking. Its purpose is to allow algorithms and sensor configurations to be physically tested and improved. Since this vehicle platform has a digital twin within the simulation environment, one can test and compare the same algorithms and autonomy stack in simulation and reality. This platform has been built with an eye towards characterizing and managing the simulation-to-reality gap. Herein, we describe how this platform is set up, deployed, and used to improve autonomy for mobility applications.
\end{abstract}

\section{INTRODUCTION}
\label{sec:intro}

\subsection{Motivation}
\label{sec:motivation}
This contribution is concerned with the idea of using simulation to facilitate the task of developing algorithms that improve the autonomy of wheeled and tracked robots operating on rigid or deformable terrains. The value proposition is tied to the cost-effective manner in which data can be generated in simulation as well as to the ease and safety with which candidate solutions can be tested and iterated upon. Using simulation in robotics requires a dynamics engine and a model \cite{karenSimRobotics2020}. 
Should one make the necessary investment to understand both how the simulator and model should be configured, simulation provides insights that are difficult to obtain in physical testing, e.g., complete state information and quantitative insights about the interaction between the robot and environment it operates in. Unfortunately, these insights do not always lead to decisions that work well in reality due to the simulation-to-reality gap \cite{sim2realGapEssex1995}. How to close this sim-to-real gap remains an open problem, and solutions have been proposed that include randomizing the experience of the robot inside the simulator \cite{domainRandomizationAbbeel2017,sim2Real2018}, using adversarial learning to capture unknown components of the model via ghost external perturbations in an adversarial reinforcement learning framework \cite{pintoAdversRL2017}, using ensembles of models \cite{levineEPOpt2016}, using a mix of simulation-generated and real-world data to train robots \cite{farchyStone2013,foxSim2RealClosing2019}, etc.

Despite these and other similarly valuable contributions, the community lacks an objective understanding of what exactly produces the sim-to-real gap. Our contribution is motivated by this observation and inspired by the belief that an open source autonomy research testbed can be a catalyst for research in two areas: improvement of algorithms for autonomy in mobility; and understanding and mitigating the sim-to-real gap. To that end, we describe herein an infrastructure, i.e., a physical and software platform, that allows one to experiment with an autonomy algorithm, e.g. a sensor-fusion approach, and have the opportunity to do so first in simulation and then quickly on the actual robot. 
The principle is to have \textit{one} ROS 2 \cite{ROS-2009,ROS2} autonomy stack that uses the same collection of perception, state estimation, planning, and control algorithms both in simulation \textit{and} in the real world. This can enable one to perform controlled tests to understand and evaluate the performance of the algorithms. 

\subsection{Contribution}
\label{sec:contribution}
\begin{figure}[t]
	\centering
	\includegraphics[width=\linewidth]{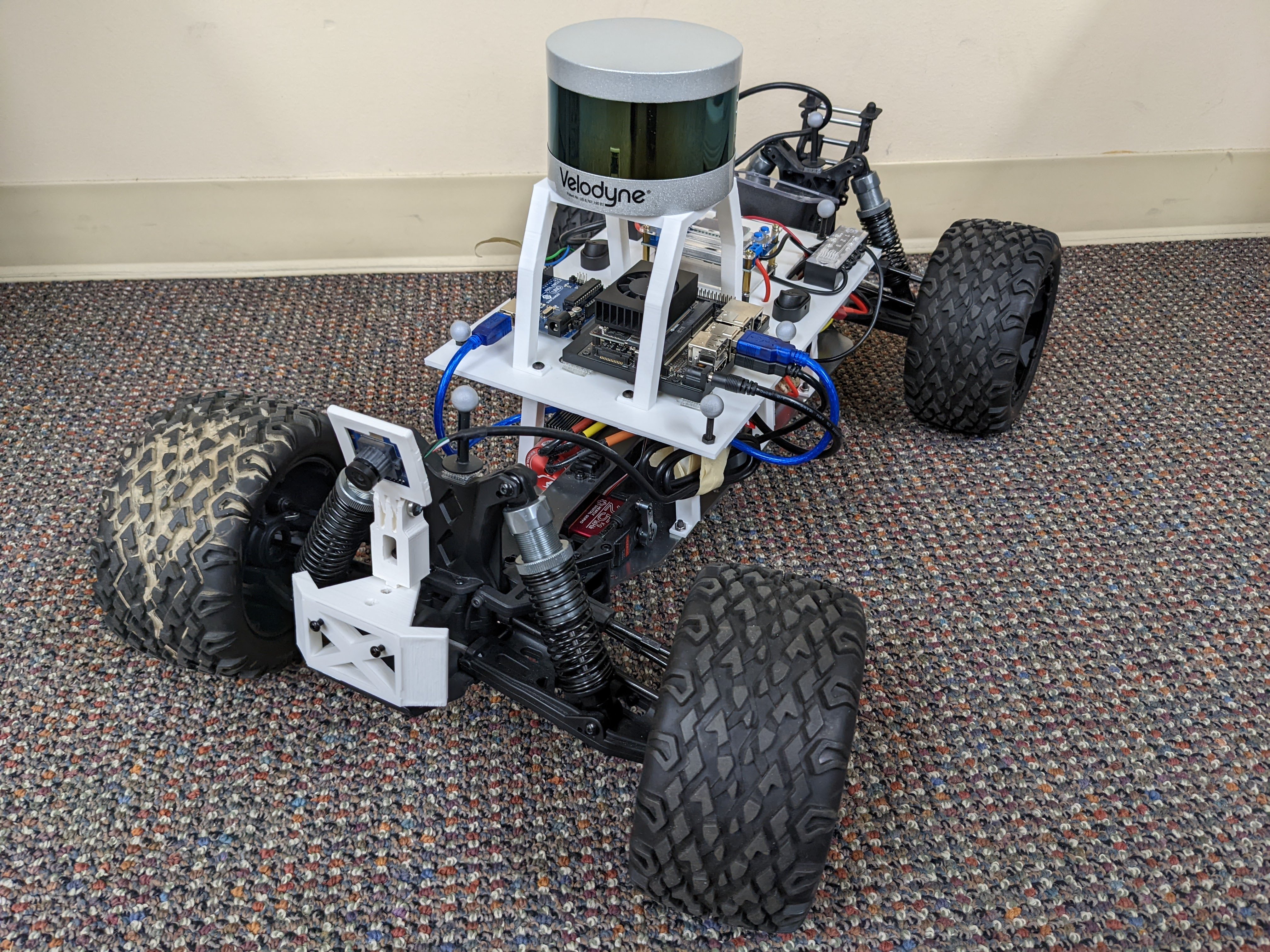}
	\caption{Fully equipped ART 1/6th scale autonomous vehicle showing 3D printed reconfigurable mountings and support of camera, 3D lidar, and Jetson Xavier NX, the later running the autonomy stack.}
	\label{fig:art_vehicle}
\end{figure}
Our contribution is twofold. First, we established an autonomy development environment and an accompanying hardware platform (called ART, from ``autonomy research testbed'') that in tandem facilitate two types of technical pursuits: research into algorithms for autonomy in mobility, in the context of wheeled and tracked vehicles; and quantitative characterization of the sim-to-real gap in robotics. Second, we developed an autonomy toolkit (ATK) which is a set of command line tools for building the container system that underpins ART.

The autonomy research testbed is containerized \cite{dockerContainers2014} to provide an OS agnostic platform that fosters collaboration and accelerates the research effort since it: 1) mitigates the startup time for setting up an autonomy stack by providing its bridges and interfaces to other system components, e.g., sensors and bridge to simulation; 2) abstracts the components of the autonomy stack thus enabling researchers to quickly replace the algorithm(s) it embeds; 3) makes available a proven simulation environment for tracked/wheeled vehicles, operating on/off-road conditions by allowing communication with Chrono \cite{projectChronoWebSite}; and, 4) leverages a 1/6th scale vehicle for real world testing that anchors the algorithm validation exercises as well as the sim-to-real inquiry efforts. The hardware component of ART (the 1/6th scale vehicle shown in Fig.~\ref{fig:art_vehicle}) has been chosen for its topology, payload capacity, and off-road ability. In term of its topology, the suspension and steering mechanism closely mirror those of a full-size vehicle. The scaled vehicle's topology and payload capacity ensure that it can host the same class of sensors encountered on a full sized-vehicle, e.g. Velodyne VLP-16. Finally, with a wider wheel base and tire format, the vehicle can operate in off-road conditions.

From a high vantage point, our main accomplishment is that a basic autonomy stack has been set up and Docker-containerized. It can be used either with the associated vehicle, to test autonomy algorithms, or in a simulator using the Chrono digital twin. This is the core of ART. Additionally, ATK provides the tools to build the ART ecosystem and collect, visualize, and analyze data coming from ART, which others can leverage for custom containerized environments. ART and ATK allow one to (a) carry out research in the sim-to-real gap; (b) generate data for data-driven solutions in robotics; and (c) improve autonomy algorithms (perception, state estimation, planning, controls) in plug-and-play fashion by working exclusively on the ART vehicle, or by combining physical testing and physics-based simulation. The simulator which ART leverages, 
has the following features: easy wheeled/tracked vehicle model set up via templates \cite{ChronoVehicle2019}; sensor simulation \cite{asherChronoSensor2021}; terradynamics support \cite{chronoSCM2019,weiTracCtrl2022}; support for multiple agents \cite{synchrono2020}; Python bindings \cite{pyChronoCondaWebSite}.

The subsequent portions of the contribution are organized as follows. To contextualize this contribution, we provide next an overview of similar community efforts. Section \ref{sec:atk} highlights the container build tools associated with this project. Section \ref{sec:art} describes the autonomy research testbed by outlining the autonomy stack and its structure, the bridge to the simulator \cite{chronoOverview2016}, the companion vehicle platform, and its digital twin. This information, along with the online documentation referenced herein should be sufficient for an interesting party to replicate and deploy ART at a modest cost should one have access to a 3D printer.

\subsection{State of the Art}
MuSHR \cite{srinivasa2019mushr}, an open-source race-car project, comes closest in spirit to ART. For simulation, MuSHR draws on Gazebo \cite{gazebo}. It uses a 1/10th form-factor vehicle, has one 2D lidar sensor, one stereo camera, and a Jetson Nano processor. MuSHR is ROS-based and runs off a Linux Ubuntu distribution. It has comprehensive documentation and has been used in classes at University of Washington. Another well known platform is MIT RACECAR \cite{racecarMIT2022,racecarMIT-github2022}. It uses a 1/10th form factor, ZED stereo camera, 2D lidar. The software stack, which is ROS-anchored, is provided as a Docker image. Simulation support comes via Gazebo. MuSHR and RACECAR share the same pros and cons -- open source platform, ROS-backend, proven software stack; and, respectively, small form factor, indoors use, software and hardware infrastructure relatively difficult to expand beyond the current vehicles. NVIDIA's JetRacer \cite{jetRacerNVIDIA2022} is a hobbyist platform that comes in a small form factor (1/18th); it uses a Raspberry Pi Camera V2, and runs off a Jetson Nano processor. This is not necessarily for use as an autonomy research platform, instead it draws on NVIDIA's Jetpack, which is a basic toolkit in Python. 

The Cat \cite{catPlatform2022} uses only IR sensors and is designed to chase an IR LED mounted on a remote control car. It has a 1/10th format, two IR sensors, a PIC18F2455 microcontroller; SSC-32 servo controller. It has a MATLAB simulator customized to the hardware platform at hand (changing hardware platform requires re-write of the simulator); no other vehicles or sensors can be exercised in the simulator. Karr \cite{karrPlatform2022} is designed to only use a depth camera, and requires walls on either side of the car to guide the vehicle along the path. It is a 1/10th form factor, uses NVIDIA's Jetson TX1 chip that comes with a developer kit targeting visual computing running in a pre-flashed Linux environment. Karr relies on ROS and trains a neural net using Keras with a Theano backend using real-life videos. There is no simulator bundled, yet Carla \cite{carlaAVsim2017} is mentioned as a next logical step towards the goal of training in simulation. 

Donkey Car \cite{donkeyCar2022} is a popular hobbyist project, which is concerned with racing small, 1/16th or 1/10th, format vehicles that mostly draw on Raspberry Pi 3b+ and a wide-angle camera. It can be fitted with Jetson Nano or TX2 with access to the NVIDIA developer kit and its ecosystem. Donkeycar uses Python for controlling the car, and Unity \cite{unityGaming} as a simulator, the latter building off the NVIDIA PhysX simulation engine \cite{physxNVIDIA}. The Donkeycar infrastructure is popular, see, for instance \cite{offshootDonkey2021} for an offshoot. One project that stands out in terms of form factor is PARV/MPAD; at 1/8th is larger than all the rest, but smaller than the platform discussed herein. While the project has modest goals on the autonomy component, it is remarkable in that it is a 3D printed hardware platform \cite{parvWPI2021} that comes with a basic autonomy stack called MPAD (Modular Package for Autonomous Driving), which is Python based, uses OpenCV with a Rasberry Pi camera, has Ultrasonic sensors, an IMU, and runs off a Rasberry Pi 4 \cite{mpadWPI2020}. Vehicle control draws on an Arduino Mega. A Raspberry Pi-based solution with Arduino for vehicle control is described in \cite{smallSelfDrivingRCCar2021}; it is small form factor and works by neural-net-enabled image recognition in conjunction with scaled-down traffic signs (data not collected in simulation, neural net trained with data collected by driving car around manually). Another hobbyist solution is PiCar of SunFounder \cite{sunFounderPiCar2022}, which at less than 1/10th format is an inexpensive platform that uses Raspberry Pi 4 and a wide-angle USB camera, with controls designed in EzBlock \cite{ezBlock2022}, an open platform for building intro-level electronic projects.

All solutions mentioned above are different in at least one of the following four aspects: (1) project breadth of scope; (2) scale and flexibility of hardware platform; (3) degree to which project embraces and promotes a simulation component; (4) documentation and support. For (1), ART is a researcher's (instead of hobbyist's) platform that uses, but is not limited to, the vehicle shown in Fig.~\ref{fig:art_vehicle}. Of the solutions mentioned above, only MuSHR has a similar research purpose. For (2), we adopted a larger form factor since the interest is on- and off-road mobility. Moreover, the goal is enabling sensor fusion, experimenting with various sensor packages, etc., which requires additional payload capacity. For (3), this group is interested in simulation first and foremost, and the vehicle is not the goal, but rather a means to an end. The ends are enabling research in autonomy for mobility and better understanding and management of the sim-to-real gap. For (4), a documentation infrastructure details our solution \cite{atk-art2022}. 

Finally, this effort does not seek to establish our autonomy testbed as a competitor to full-blown stacks produced by companies such as Tesla, Waymo, or Motional, or made available in Autoware \cite{autoware2018} or Apollo \cite{apolloAStack2022}. Our goal is to enable easy, plug-and-play research, that enlists the support of a basic autonomy stack and a companion platform for simulation-enabled research in autonomy. The research facilitated by ATK and ART pertains to advancing the state of the art in relation to what simulation can and should do in robotics; the physical platform provides the required reality check and ground truth.

\section{AUTONOMY TOOLKIT}
\label{sec:atk}


The objective of ATK is to provide a modular and portable framework for developing, testing, and deploying autonomous algorithms in simulation and reality. This toolkit is a Python package that leverages Docker by wrapping Docker Compose to build and deploy a multi-container system specifically designed for autonomy research. ATK wraps Docker Compose in that all of the functionality of Compose is still available; however, ATK also provides utilities, defaults, documentation, and examples specifically for applications relating to autonomous algorithm development. ATK is open-source, available on GitHub, and has been made available through the Python Package Index (PyPI) \cite{atk-pypi}.

A requirement of this toolkit is to be OS portable. To that end, a containerized system, set up to leverage Docker, is separated into ``services'' that can be combined to produce a container network that supports complex interactions across a variety of projects. To expedite the deployment process to the physical hardware under test, the same containerized system is used. The Docker interface, especially on Linux systems, is extremely lightweight and introduces negligible overhead for most applications. The development framework is built into two main components: the ATK Python package and the Docker containers themselves. The ATK package, which helps generate the containers, is described in Section \ref{section:autonomy-toolkit-package}; the container system is described in Section \ref{section:container-system}.

\subsection{autonomy-toolkit Package}
\label{section:autonomy-toolkit-package}

The package which generates the container system is, in its simplest form, a wrapper of Docker Compose. Called \textit{autonomy-toolkit}, \textit{ATK} for short, this Python package is publicly available and provides a cross-platform command-line interface. ATK leverages the multi-container architecture to provide optional, extensible, and customizable containers to facilitate development of AVs that greatly increases the generality of the development workflow. The only non-Python requirement for the autonomy-toolkit is Docker. With Docker available, the toolkit can be installed through typical python mechanisms like \texttt{pip}. ATK works as follows: YAML configuration files are combined with defaults provided through ATK into a configuration file readable by Docker Compose. From there, ATK simply makes calls to Docker Compose based on the resulting output. All defaults provided through ATK are customizable. Furthermore, the generic Docker Compose commands can be used in conjunction with the generated config file, if desired, without ATK.

ATK was built with modularity and expandability in mind. This means the functionality of the toolkit itself is agnostic of the autonomy stack it is used for, i.e. it is not tied to ART. Individual hardware platforms and control stacks may implement their own containers and customize the default configurations, provided they follow the few requirements outlined in the documentation \cite{atk-art2022}. As ATK is agnostic of the hardware platform and usage, it can also be used to generate custom containers outside the original scope of the toolkit. For instance, the simulation container to be discussed further in Section \ref{sec:simulation} leverages ATK to create its own container for Chrono. Other common use cases could include a container for training data or an HD map, a deployment container for specific hardware, an automated testing environment, etc.
	
\subsection{Container System}
\label{section:container-system}

Distributed with the ATK package itself are many predefined and customizable utilities that are used to generate the images and containers. In general, there are two primary services that come directly with the ATK package: \texttt{dev} and \texttt{vnc}. 

	\textbf{\texttt{dev}:} This is the primary component that is used for algorithm development. It is defined by a custom Dockerfile which utilizes build arguments specified through a configuration file to generate an image specific to the project being developed. By default, \texttt{dev} builds on the ROS 2 Galactic image, the most recent ROS 2 distribution at the time of writing. By default, \texttt{dev} is assumed to be the primary container and, when launched, initializes a shell environment. The directory that holds the configuration file is mounted into the container for data persistence upon container termination. Other utilities and configuration are preformed to further enhance the generality of the toolkit and remove OS dependence.

	\textbf{\texttt{vnc}:} Since the isolation-based design of Docker results in difficulty visualizing applications, a cross-platform visualization system is used which requires no host setup. 
	The implemented solution leverages Virtual Network Computing (VNC) within a separate service, making its usage optional.
	To use VNC visualization, \texttt{dev} (or any other container run using ATK) must simply attach to the same network and configure its \texttt{DISPLAY} variable to that of the \texttt{vnc} hostname.
	NoVNC, a browser based VNC client, can then be used to view any displayed windows from \texttt{dev} from any internet browser. Further explanation of the use, and a short tutorial are provided with the ATK documentation \cite{atk-art2022}.

Note that a networking layer is set up to facilitate distributed container interaction. This feature supports hardware-in-the-loop experiments, where the simulation container is deployed on separate hardware from the autonomy stack.
To facilitate this, we leverage Docker and Docker Compose networking features. When services are deployed on the same system, the default network configurations are used. However, when the services are deployed in a distributed fashion, the network is customized to expose the necessary ports to facilitate inter-computer communication.

\begin{figure}[t]
	\centering
	\includegraphics[width=\linewidth]{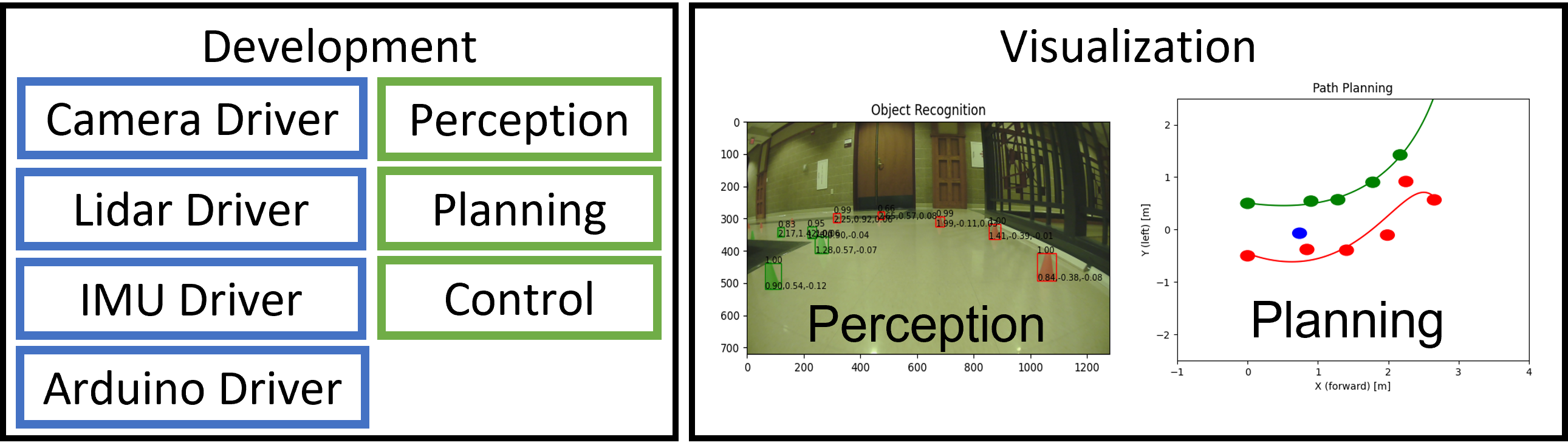}
	\caption{ART environment, setup used on the physical vehicle.}
	\label{fig:dev_env_real}
\end{figure}

\begin{figure}[t]
	\centering
	\includegraphics[width=\linewidth]{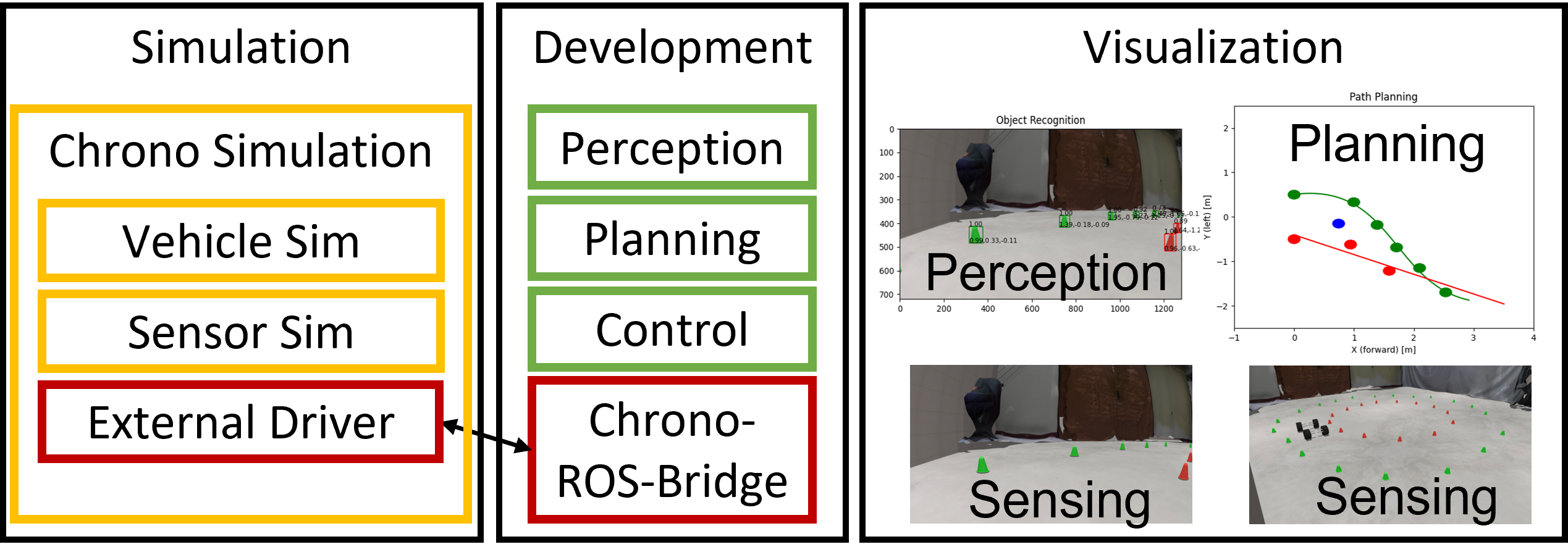}
	\caption{ART environment, setup used for simulation.}
	\label{fig:dev_env_sim}
\end{figure}

Since the environment is containerized, it can be deployed to a physical vehicle, with optional hardware-specific optimizations. While ATK is general purpose, an example of the containerized system can be seen in Figs. \ref{fig:dev_env_real} and \ref{fig:dev_env_sim}, which illustrate the services when running on the real vehicle vs. simulation, respectively. 

\section{AUTONOMY RESEARCH TESTBED}
\label{sec:art}

\subsection{Simulation}
\label{sec:simulation}

One of the motivations behind this effort was the use of simulation as a means of designing and testing autonomous algorithms. To support this, a bridge to Chrono \cite{chronoOverview2016} was developed to allow direct integration within the containerized system for ART. The bridge, along with the custom Chrono container, is an example of the modular environment produced by ATK. 

The Chrono ROS Bridge is a ROS 2 package that integrates with Chrono to feed messages to and from the simulation scenarios. The interface is built on JSON messages so users can send \textit{any} data between ROS and Chrono. Generic publishers and subscribers (a feature of ROS 2) are leveraged to allow for user defined message types and topic names to be used. The package, written in C++, is open source and available under a BSD3 license. 

On the Chrono side, the bridge builds directly on core functionality available through the utility class ChSocket. Additional hooks were provided that wrap the JSON generation from user code. This was done to facilitate Python wrapping with PyChrono, the Python bindings to the C++ Chrono API. This allows the researchers to leverage the rapid development process enabled by Python. Additionally, to allow custom message types to be sent between ROS and Chrono, custom functors may be implemented to generate and/or parse custom message formats.

\subsection{Autonomy Stack}
The autonomy stack developed for ART is built on top of ROS 2 and includes basic implementations of autonomy algorithms to enable autonomy experiments in simulation and reality. The stack is basic; there is nothing particularly novel in its implementation. It utilizes publicly available algorithms and packages typically shipped with ROS 2. This is because the focus of ART is not on the autonomy stack itself, but the extent to which the algorithms are transferable between sim and reality. Against this backdrop, the perception algorithm is a custom trained instance of Faster-RCNN \cite{ren2016faster} built on a MobileNetV3 \cite{mobileNetV3} network. MobileNet is intended for mobile phone CPUs, but was adopted for ART considering the limited compute power available on the vehicle. For the example used later in this contribution, in which a vehicle navigates down a lane set up with cones, the perception algorithm was trained using both simulated and real images. Based on the bounding box output from Faster-RCNN, 3D cone positions are generated at each timestep and a simple planner and controller are used to keep the vehicle on the path defined by the cones. To control the car, throttle, braking, and steering inputs are communicated to the car via an Arduino processor. ROS 2 stock interface packages are utilized to communicate with sensors.

\subsection{Vehicle Chassis}
The vehicle platform is built upon a 1/6th scale remote controlled car. With a 47 cm wheel base and a 34 cm track width, the vehicle is large enough to carry multiple cameras and an automotive lidar such as the VLP-16 at 0.83 kg. This is critical since research into the difference between simulation and reality when applied to sensor simulation must perform analysis on the specific sensors of interest in full-scale applications. The base vehicle includes a double wishbone independent suspension at the front and rear. The kinematics of this suspension allow for use in off-road mobility. For actuation, the vehicle includes a 1300 KV brushless motor and a 15 kg-cm servo for steering which we upgraded to a 25 kg-cm servo for durability. The servo controls the steering through a Pitman arm steering mechanism.

\subsection{Electronic Hardware}
The base RC car was augmented to allow for direct control of the steering and throttle via an onboard computer. The modifications were minimal to allow for duplication of the system. First, the dual-battery setup powering the electronic speed controller (ESC) was reduced to a single battery, which with an eye towards safety, lowers the top speed of what would otherwise be a very fast car. The use of a dedicated battery for electronics ensures that voltage spikes and drops induced by the motor do not affect any sensitive compute or sensing electronics. To prevent floating voltage issues, all ground connections are wired to a common ground rail. A power distribution and control wiring diagram along with a bill of material are provided with the hardware documentation online \cite{atk-art2022}.


The computer used on the ART vehicle is a Jetson Xavier NX, although other compute platforms could also be used. Indeed, the design of the vehicle and development environment are compute system agnostic. To facilitate this, direct control of the servo and ESC is performed through PWM signals from an Arduino chip. On the vehicle herein, an Arduino Uno was leveraged, but an Arduino Nano is considered in the documentation. The Arduino is used exclusively as a device driver, performing no autonomy. The ART vehicle is equipped with a USB camera and a VLP-16 lidar. Further sensors can be mounted on the vehicle to facilitate additional research needs.

\subsection{Mounting, Expansion, and Reconfigurability}
The ability to reconfigure the vehicle platform was a high priority given the ``research-platform mission'' this solution must fulfill. To that end, all mounting components were designed to be 3D printed, with a base plate for electronics that could be laser cut or 3D printed. The final setup is shown in Fig. \ref{fig:art_vehicle} with the electronics mounted above the motor and ESC on the base RC car. While most electronics are mounted directly to the electronics board, the camera and lidar must be mounted in unobstructed locations. To achieve this, a custom mount allows a lidar to be placed above and clear of the rest of the vehicle, and a front 3D printed bumper allows for camera mounting. Future mountings will be designed to hold multiple cameras and tracking markers on the front and rear bumpers.  In addition to electronics and sensors, motion capture tracking targets can be mounted to the car using a scattering of holes across the vehicle.

\subsection{Vehicle Digital Twin}
\begin{figure}[ht]
	\centering
	\begin{subfigure}{.49\linewidth}
		\centering
		\includegraphics[width=\linewidth]{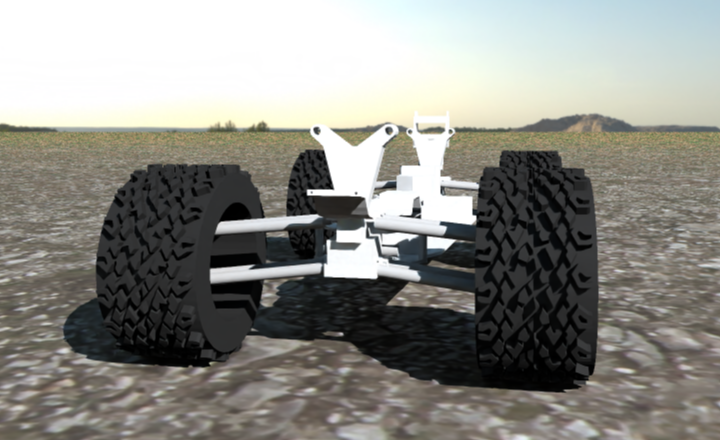}
		\caption{Front view of ART model.}
		\label{fig:chrono_model_front}
	\end{subfigure}%
	\hspace{.01cm}
	\begin{subfigure}{.49\linewidth}
		\centering
		\includegraphics[width=\linewidth]{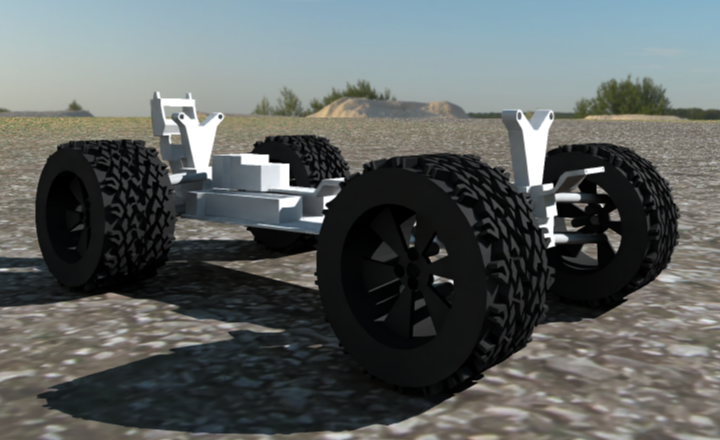}
		\caption{ISO view of ART model.}
		\label{fig:chrono_model_iso}
	\end{subfigure}
	\caption{Chrono::Vehicle model of ART as visualized with Chrono::Sensor.}
	\label{fig:chrono_model}
\end{figure}

The vehicle described in this contribution has a digital counterpart modeled in Chrono using Chrono::Vehicle \cite{ChronoVehicle2019} and Chrono::Sensor \cite{asherChronoSensor2021}. The vehicle model is implemented with a double wishbone suspension and a linear spring-damper with parameters estimated from the vehicle mass. Further calibration and measurements will allow for more accurate values of spring stiffness and damping coefficient. The Chrono model of the car is rendered in Fig. \ref{fig:chrono_model} using Chrono::Sensor and highlights the mesh representation of the car along with the double wishbone configuration.

The steering model uses the Chrono::Vehicle Pitman arm template, allowing actuation of the steering arm. The maximum steering angle was calibrated from a set of minimum radius turn tests using motion tracking. The motor model is a simple linear torque-speed curve, with decreasing power with motor speed. While this does not accurately model the brushless motor in a standalone configuration, a lumped model of the ESC and motor must be considered. Calibration of the motor model remains part of future work and will be critical as more dynamic simulations come into focus.

\section{DEMONSTRATION}
\label{sec:demo}

\begin{figure}[ht]
	\centering
	\begin{subfigure}{\linewidth}
		\centering
		\includegraphics[width=\linewidth]{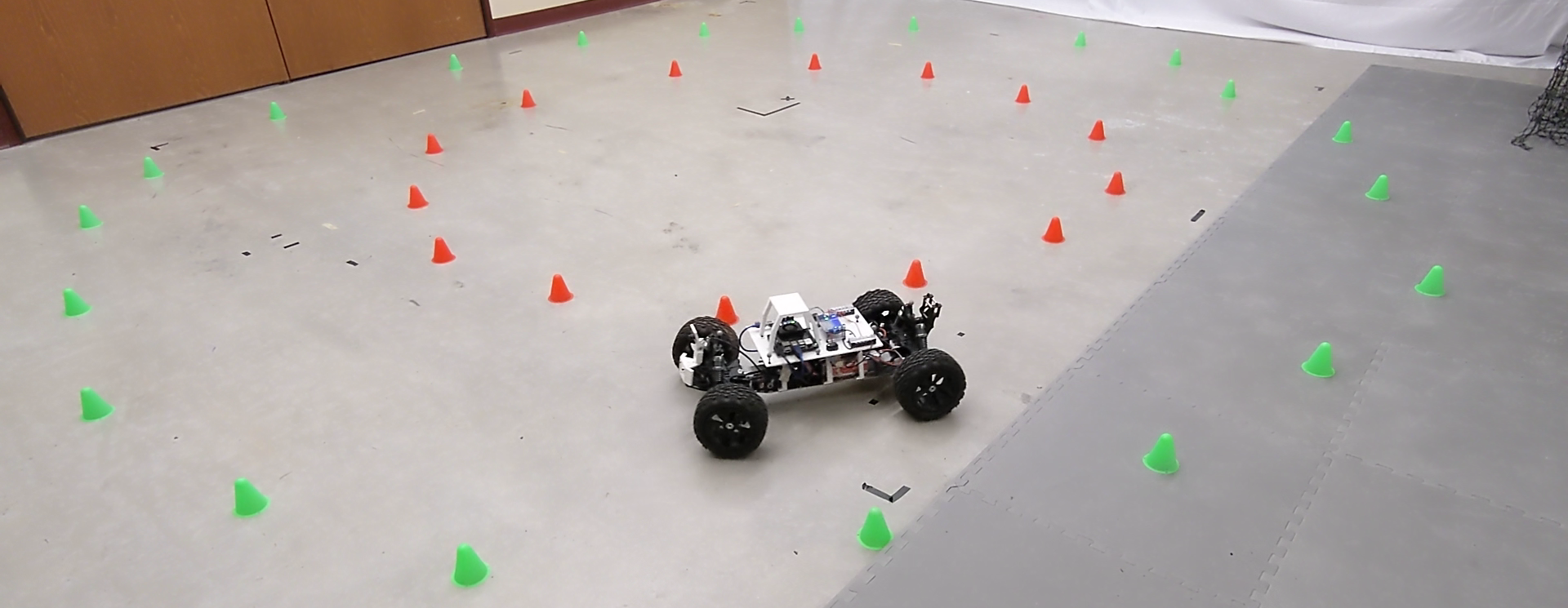}
		\caption{Real setup}
		\label{fig:real_3rdperson}
	\end{subfigure}
	\begin{subfigure}{\linewidth}
		\centering
		\includegraphics[width=\linewidth]{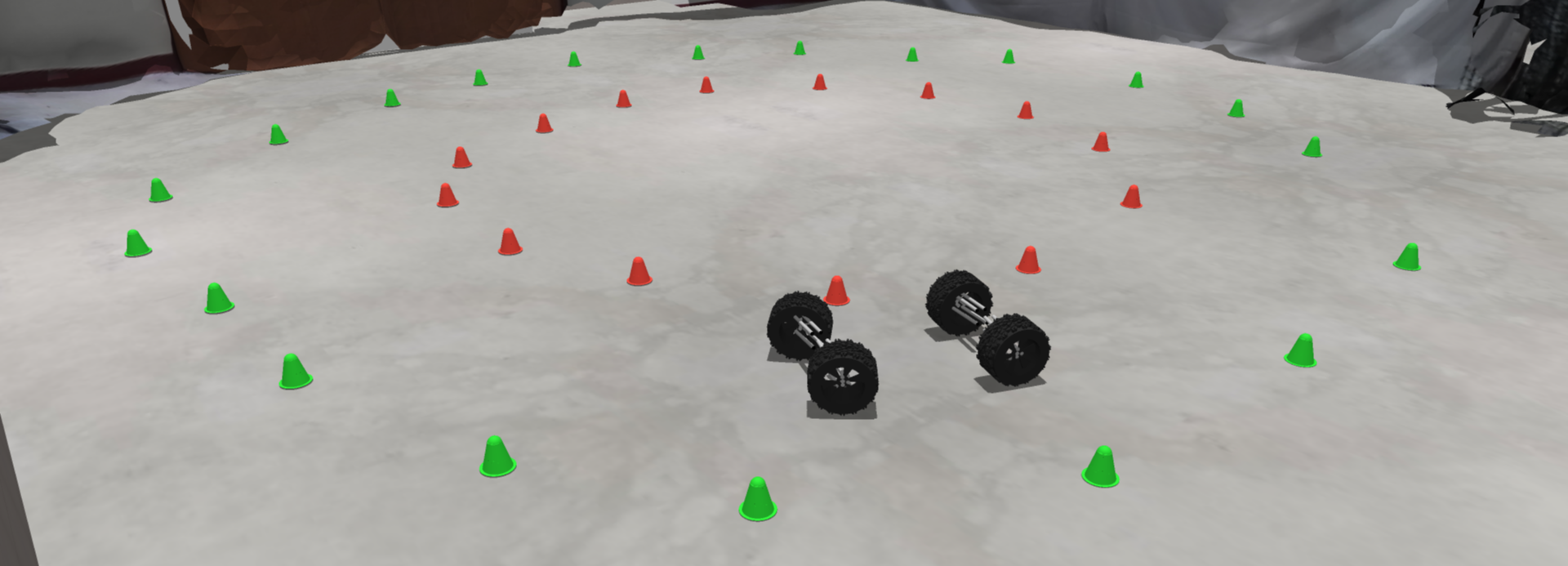}
		\caption{Simulated setup}
		\label{fig:sim_3rdperson}
	\end{subfigure}
	\caption{Real and simulated scenarios using same cone locations measured by a motion capture system.}
	\label{fig:sim_real_1_to_1}
\end{figure}

To demonstrate ART at work, a path of cones was setup in our motion tracking lab. The cone locations were recorded and injected into a simulation of the same motion. The vehicle then drove along the path in simulation and reality. Images of this setup from a similar location for real and sim are provided in Fig. \ref{fig:sim_real_1_to_1}, which shows the same cone paths and the vehicle in a similar location on the path. To navigate, the vehicle exclusively uses its front facing camera. This demonstration is included in the supporting video and can be found online \cite{art-iros-video}.

\begin{figure}[ht]
	\centering
	\begin{subfigure}{\linewidth}
		\centering
		\includegraphics[width=\linewidth]{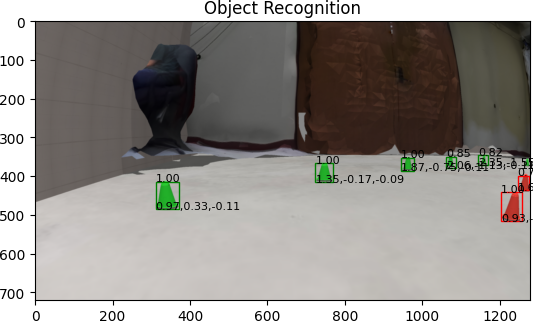}
		\caption{Perception on a simulated image}
	\end{subfigure}
	\begin{subfigure}{\linewidth}
		\centering
		\includegraphics[width=\linewidth]{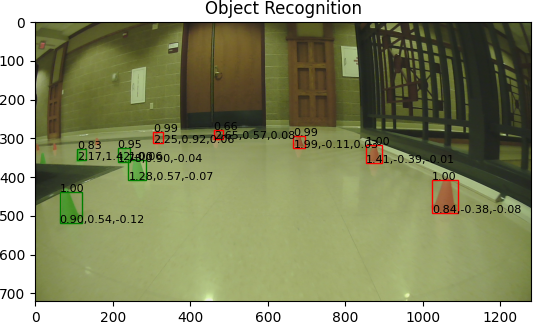}
		\caption{Perception on a real image}
	\end{subfigure}%
	\caption{Intermediate perception results from the autonomy stack shown for simulation and reality. Overlaid on the image are the detected bounding boxes, classes, confidence, and estimated 3D location relative to the vehicle.}
	\label{fig:autonomy_stack_perception}
\end{figure}

\begin{figure}[ht]
	\centering
	\begin{subfigure}{\linewidth}
		\centering
		\includegraphics[width=\linewidth]{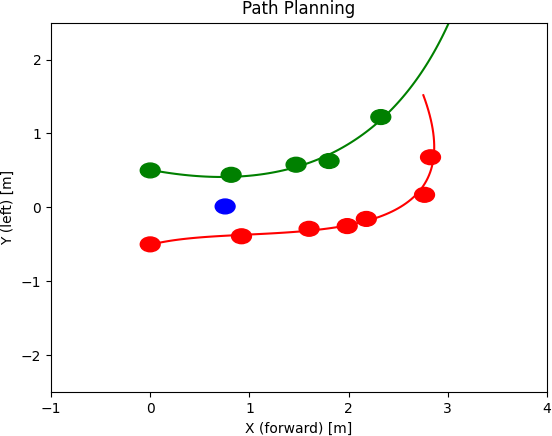}
	\end{subfigure}
	\caption{An example of intermediate results from the planning stage. Shown here are the locations of the cones in 2D (circles), the estimated curve of the boundary, and the target point used for control (blue).}
	\label{fig:autonomy_stack_planning}
\end{figure}

To further demonstrate the algorithms used by ART, we show examples of the intermediate output from the perception and planning stages. Perception results from simulation and reality are show in Fig. \ref{fig:autonomy_stack_perception}, and results from the planning stage are shown in Fig. \ref{fig:autonomy_stack_planning}.

\section{CONCLUSION AND FUTURE WORK}
\label{sec:conclusion}

This contribution outlines an open-source autonomy research testbed (ART) whose purpose is twofold: conduct research in autonomy for wheeled/tracked vehicles, in on/off-road conditions; and, investigate the sim-to-real gap in robotics -- understand what causes it, and how it can be controlled. Looking ahead, we will equip ART with additional sensors (e.g. GPS, IMU, and extra cameras including a stereo camera). Given that we can track the position of the vehicle in reality with millimeter accuracy via a motion capture system, this richer family of sensors will allow us to investigate more expeditiously better sensor models, as well as perception, state estimation, planning, and controls algorithms, and understand how inaccuracy in different components of the autonomy stack propagate downstream and cause the sim-to-real gap. Another future direction is tied to setting up ART to work with tracked vehicles and in off-road conditions.

Finally, ATK can be used to facilitate autonomy algorithm development by allowing researchers to configure a custom set of containers to host their development, deployment, simulation, and visualization needs. For instance, this platform will assist UW-Madison in their SAE-sponsored AutoDrive Challenge II.





\section*{ACKNOWLEDGMENT}
This work was carried out in part with support from National Science Foundation project CPS1739869. Special thanks to the Society of Automotive Engineers (SAE) and General Motors for their support through the AutoDrive Challenge Series II competition.


\bibliographystyle{ieeetr}
\bibliography{BibFiles/refsSensors,BibFiles/refsMachineLearning,BibFiles/refsAutonomousVehicles,BibFiles/refsChronoSpecific,BibFiles/refsSBELspecific,BibFiles/refsMBS,BibFiles/refsCompSci,BibFiles/refsTerramech,BibFiles/refsFSI,BibFiles/refsRobotics,BibFiles/refsDEM}

\end{document}